\documentclass{article}
\usepackage[nonatbib,preprint]{neurips_2020}
\usepackage{etoolbox}
\usepackage[title]{appendix}
\makeatletter

\usepackage[utf8]{inputenc} 
\usepackage[T1]{fontenc}    
\usepackage{hyperref}       
\usepackage{url}            
\usepackage{booktabs}       
\usepackage{amsfonts}       
\usepackage{nicefrac}       
\usepackage{microtype}      
\usepackage{makecell}
\usepackage{multirow}
\usepackage{microtype}
\usepackage{array}
\usepackage[sort, numbers]{natbib}
\usepackage{mathtools}
\usepackage{mathrsfs}
\usepackage{subfigure}
\usepackage{booktabs} 
\usepackage{algorithmic}
\usepackage{algorithm}
\usepackage{amsmath}
\usepackage{xparse}
\usepackage{xcolor}
\usepackage{hyperref} 
\usepackage{makecell}
\usepackage{multicol}
\usepackage[export]{adjustbox}
\usepackage{multirow}
\usepackage{tikz}
\usetikzlibrary{bayesnet}
\usetikzlibrary{arrows}
\usepackage{color}
\usepackage{graphicx}
\usepackage{caption}
\usetikzlibrary{backgrounds}
\usepackage{bm}
\hypersetup{
     colorlinks   = true,
     citecolor    = green
}

\graphicspath{{plots/}}
\usepackage{enumitem}

\title{Attentive Feature Reuse for Multi Task Meta learning}

\author{
  Kiran Lekkala\thanks{Correspondence to Kiran Lekkala. Authors are with iLab, Department of Computer Science,University of Southern California, Watt way, Los Angeles, CA (webpage, alternative address)} \\
  Department of Computer Science\\
  University of Southern California\\
  Los Angeles, CA 90089 \\
  \texttt{klekkala@usc.edu} \\
  \And
  Laurent Itti \\
  Department of Computer Science \\
  University of Southern California \\
  Los Angeles, CA 90089 \\
  \texttt{itti@usc.edu} \\
}

\usepackage{tabularx}

\newcommand\clearrow{\global\let\rowmac\relax}
\clearrow

\begin{document}

\maketitle

\begin{abstract}
We develop new algorithms for simultaneous learning of multiple tasks (e.g., image classification, depth estimation), and for adapting to unseen task/domain distributions within those high-level tasks (e.g., different environments). First, we learn common representations underlying all tasks. We then propose an attention mechanism to dynamically specialize the network, at runtime, for each task. Our approach is based on weighting each feature map of the backbone network, based on its relevance to a particular task. To achieve this, we enable the attention module to learn task representations during training, which are used to obtain attention weights. Our method improves performance on new, previously unseen environments, and is 1.5x faster than standard existing meta learning methods using similar architectures. We highlight performance improvements for Multi-Task Meta Learning of 4 tasks (image classification, depth, vanishing point, and surface normal estimation), each over 10 to 25 test domains/environments, a result that could not be achieved with standard meta learning techniques like MAML.
\end{abstract}

\section{Introduction}
\label{sec:introduction}
Existing methods in Multi-task learning \cite{LiuJD19, MisraSGH16} leverage inter-modal features by training on multiple modalities/tasks together, where these tasks are assumed to be fixed. Here, we define {\bf task} as achieving different goals and outputs; e.g., image classification, depth estimation, or surface normal prediction. Since data is prone to domain and task shift, it is essential to consider these shifts. Meta learning \cite{RaviL17} addresses task shift by learning a set of variants for a given task (which we here define as {\bf subtasks}; e.g., learning different sets of classes or from different datasets within a general object recognition task), in such a way that the model can quickly generalize to new unseen datasets. Current methods in meta learning deal only with subtask variants having identical output dimensions and loss functions \cite{FinnAL17}, making them unusable for more heterogeneous situations.

Here we consider the problem of scaling adaptive learning by combining multi-task and meta learning. Particularly, we extend  both methods by making a model learn multiple high-level tasks/modalities concurrently, and meta-adapting to new subtasks/datasets within those modalities. As prior research suggests, we expect to derive advantages like learning better features for related tasks \cite{abs-1905-07553}, and fast, few-shot adaptation to new subtasks \cite{VinyalsBLKW16}.

Popular methods in meta learning use multiple evaluation benchmarks, ranging from image classification to pose regression. Although earlier works typically train and test within one modality (e.g., different subsets of mini ImageNet \cite{VinyalsBLKW16}), recent works extend to multiple datasets \cite{TriantafillouZD20}, yet still within a specific task (e.g., image classification). In multi-modal meta learning, \cite{YaoWHL19} recently proposed a method which  selects task-specific clusters for network parameters, while \cite{abs-1812-07172} modulate the parameters of the neural network during adaptation based on the modality of the subtask. Although these methods deal with broader data distributions, they are limited to a specific task with fixed output structure. \cite{abs-1904-09081} discuss an interesting idea of generalizing adaptation to various output structures, but only for fully-connected layers.

\begin{figure*}
\centering
  \includegraphics[width=5in]{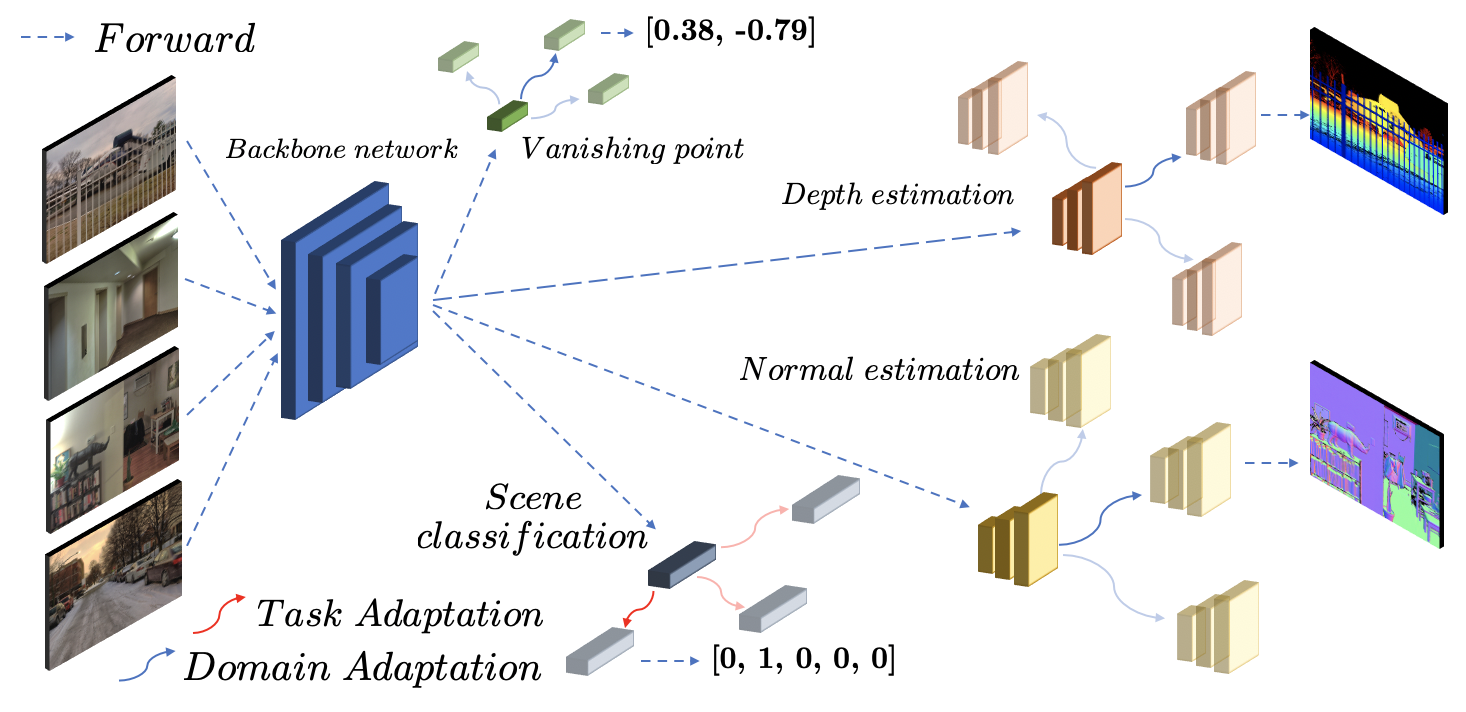}
  \caption{Proposed setting. We train a model to solve multiple high-level tasks, each of them having different output dimension, using a shared backbone network. We further include different low-level subtasks under each task. Each task-specific head is adapted to perform an unseen subtask optimally.}
  \label{fig:setting}
\end{figure*}

Our contributions are towards developing a formalism for a shared network learnt by extracting reusable feature representations from different tasks together, along with meta-heads on each task that are used to adapt to new subtask quickly. We then propose an attention mechanism, which learns to weight each filter in the backbone-network based on its relevance to a provided task (Fig.~\ref{fig:setting}). We believe this approach represents a first step in \textit{task adaptation} for multiple modalities.

\section{Related Work}

{\bf Multi Task Learning} involves learning multiple high-level tasks concurrently, and executing all of them at test-time. Existing methods  predominantly involve either using novel hard or soft shared model parameters. Since different tasks have different feature learning rates, \cite{KendallGC18} propose task loss weighting schemes, which balance the loss by enabling tasks to regularize each other during training. Recent works have proposed novel architectures to enhance multi-task learning \cite{LiuJD19, abs-2003-14058, MisraSGH16}, but all of them are geared towards training and testing on a fixed set of domain-specific tasks \cite{abs-2004-04398, GebruHF17}. 

{\bf Visual Attention} has been used by researchers in vision and language models alike. Earlier works on top-down attention for CNNs try to learn channel dependencies via a fully connected layer \cite{HuSS18}. \cite{WangGZ0D020} uses attentive feature selection and distillation for transfer learning, which was partly inspired by \cite{GaoZDMX19}. In few-shot learning, attention is used in \cite{Prol2018, HouCMSC19} to highlight features which tend to maximize the correlation between support and query samples. However, most existing approaches are limited to classification, as they use the discrete class information. Our attention method, in contrast, generalizes to any application, as we weight feature maps in the feature backbone based on a specific task.

\begin{figure}
\includegraphics[width=\textwidth]{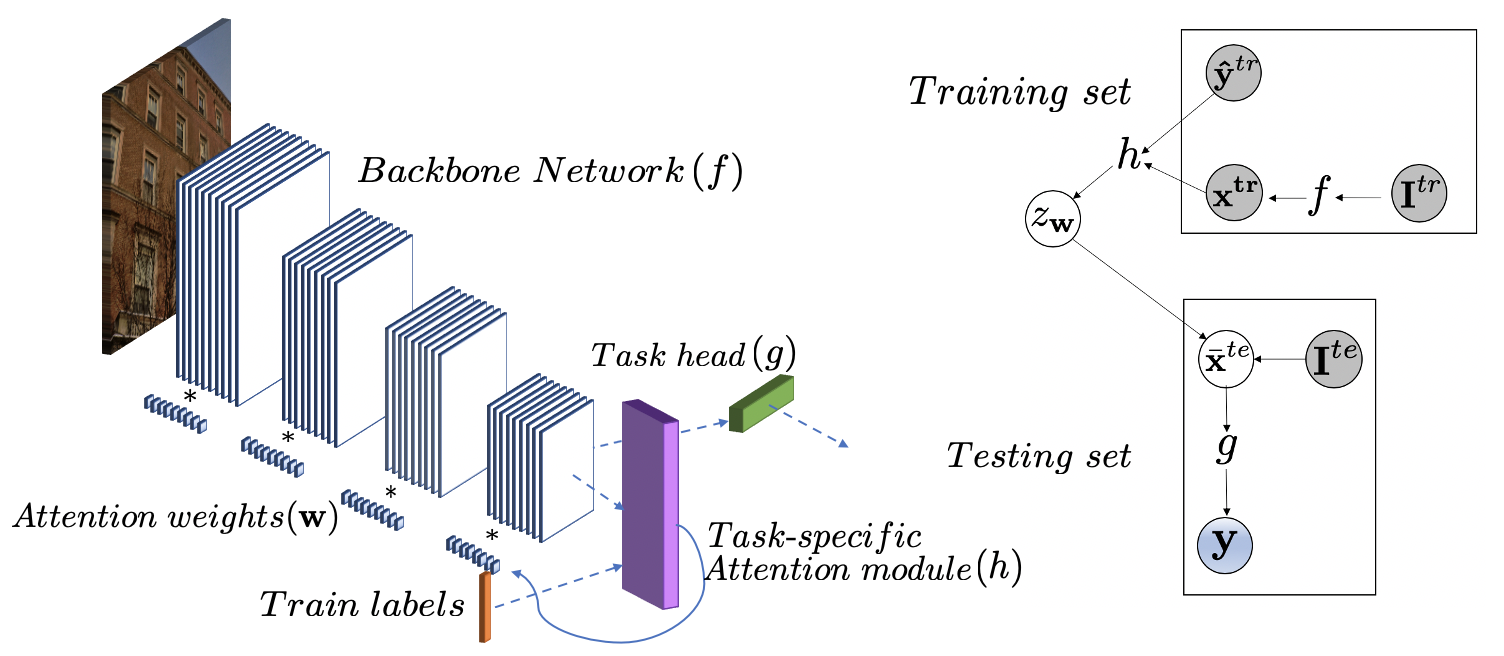}
\caption{Left: Architecture. The common backbone network learns representations for all tasks. At test time, the method needs to solve a new subtask within a fixed set of high-level tasks. Right: Graphical model of the system. Shaded and unshaded circles indicate observed and hidden variables respectively. Blue circle indicates that the target of the variable is observed during training and hidden during inference. See Section 3 for notations.}
\label{fig:arch}
\end{figure}

{\bf Meta Learning}  deals with applying prior knowledge from various tasks to learn a new task in a few shot setting (note: although prior works mention "tasks", these so far have been subtasks per our terminology). One of the most promising methods (MAML) is optimization-based \cite{FinnAL17}. During meta-training, MAML learns a parameter initialization which enables to model to quickly adapt to a new unseen subtask in a few shot setting. This involves computing Hessian-vector products which introduce computational instabilities. To alleviate these problems and scale meta learning, there have been many improvements \cite{RajeswaranFKL19}. In some applications like classification, other categories of meta learning algorithms, namely black-box \cite{BertinettoHTV19} and parametric methods \cite{SnellSZ17} also achieve state of the art results. We limit our discussion to optimization-based methods as we are concerned with flexible meta learning involving various heterogeneous tasks with varied output structures and loss functions \cite{Hospedales}. Meta learning is also applied to Domain adaptation as seen in \cite{KangF18}.

Lately, works aim to explain the effectiveness of Meta learning approaches concerning representation and adaptation aspects \cite{RaghuRBV20}. Their findings surprisingly indicate that the success behind MAML is primarily due to feature reuse amongst different subtasks. \cite{RaghuRBV20} presented an analysis of MAML, which show that actual parameter adaptation happens only in the last layer(s) and the test accuracy depends on the quality of features learnt during meta learning. In fact, several modern few-shot methods use a fixed feature learning backbone and adapt/update only the final layer during test-time \cite{abs-2003-11539}. These methods, surprisingly, beat MAML by a significant margin. Inspired by this, we follow a similar style, where we learn a fixed feature extractor, which is not adapted at all. We also include multiple heterogeneous tasks, as prior multitask learning research has shown evidence of positive transfer. Since the quality of representations is important to excel at meta-test time, we hypothesize that, better, if not equal, quality of features can be learnt by learning multiple modalities together. Furthermore, to foster task-specific feature selection, we develop a mechanism which attentively reuses features based on the tasks seen at test time. Compared to other modulation approaches \cite{PerezSVDC18}, which modulate model parameters, using our approach, the model would learn the required bottom-up features without any additional gradient flow apart from the standard backprop. In this way, we improve meta learning and create multitask meta learning, just by using a few additional parameters, as we show in our experiments.

\section{Preliminaries}
Let $ T = \{\mathscr{T}_1, \mathscr{T}_2, .. \mathscr{T}_M\}$ be a finite set of high-level tasks which a model needs to execute. A high-level task $\mathscr{T}_j$ may be specialized into sub-tasks or domains $\mathcal{T}^j_i$. Note that a separate set of sub-tasks under $\mathscr{T}_j$ is used for testing. We use a single common backbone network $f_{\Phi}$, along with a task specific head $g^j_{\theta}$, parameterized by $\Phi$ and $\theta$ respectively, to perform any given $\mathcal{T}^j_i$. $g^j_{\theta}$ denotes that $g^j$ is paramerized by $\theta_j$. For a specific $\mathcal{T}^j_i$, the model can execute it using the head $g^j_{\theta}$, when attached to backbone network $f_{\Phi}$. In other words, $f_{\Phi}$ has multiple heads and depending on the subtasks from a specific $\mathscr{T}_j$, the task-specific $g^j$ is chosen. All $\mathcal{T}^j_i$ within $\mathscr{T}_j$ share the same meta-parameters $g^j$ and adapt to $g^j_{\theta^{'}}$ to optimally perform on the given few-shot data. Final output of a task therefore is: $g_{\theta}(f_{\Phi}(\mathbf{I}))$, where $\mathbf{I}$ are the inputs to the network and their associate labels $\mathbf{\hat{y}}$.

To illustrate our notations, in most general cases, $f$ is a CNN, and $g$ is either a fully connected layer or a convolution transpose block, depending on the task. High-level tasks can be, for example, classification, depth estimation, etc. and low-level subtasks under each high-level task can be those from different domains or output objectives. For example, under classification, subtask $\mathcal{T}^1_1$ can be to classify 5 different types of fruits and $\mathcal{T}^1_2$ to classify 5 different types of vegetables. Similarly under depth estimation, $\mathcal{T}^2_1$ could be to estimate monocular depth of images taken on roads and $\mathcal{T}^2_2$ could be for indoor scenes. For each task $\mathcal{T}^j_i$, the model is trained on a train dataset associated with a task $\mathcal{D}_{train}$, which consists of task inputs and labels $\{I_u, \hat{y}_u\}_{u=0}^{U}$, and evaluated on $\mathcal{D}_{test}$ consisting of $\{I_v, \hat{y}_v\}_{v=0}^{V}$. $f_{\Phi}(\mathbf{I})$ gives out the activations of the last convolution block represented by $\mathbf{x}$ which is passed as an input to $g^j$, producing $\mathbf{y}$. $\mathbf{x}$ and $\mathbf{y}$ can have superscripts corresponding to train or test set. $\mathcal{L}_i^j(\mathbf{x}, g_{\theta}^j)$, is the loss obtained for input $\mathbf{I}$ for a specific task $\mathcal{T}_i^j$. $\mathcal{\hat{L}}_i^j$ is the train loss of a task, which is used for head adaptation and test loss $\mathcal{L}_i^j$ is used for either meta-training or evaluation at test time. During training, validation and testing, all the subtasks are sampled from different task distributions for all the high-level tasks $\mathscr{T}_j$. $\alpha$, $\beta$ and $\gamma$ are the learning rates for the head adaptation, backbone and attention network respectively. Please note that our problem setting is based on the assumption that there does not exist $\Theta = \{\Phi, \theta\}$ which can be used to execute all tasks $\mathcal{T}_i^j \in T$ optimally. This assumption is reasonable in our case, as our tasks are heterogeneous with different output dimensions.

 \section{Method}
 In this section we present the method used for training using the notations of Sec.~3.

\subsection{Multi-task Meta Learning}
In standard optimization-based meta learning \cite{FinnAL17}, there is no separate body and head, as all model parameters are used for adaptation. That is, these methods minimize the objective:

\begin{equation}
\begin{aligned}
    \min_{\Phi} \sum_{\mathcal{T}_i \sim p(\mathcal{T})} \mathcal{L}_{\mathcal{T}_i}(f_{\Phi^{'}}) = \sum_{\mathcal{T}_i \sim p(\mathcal{T})} \mathcal{L}_{\mathcal{T}_i}(f_{\Phi - \alpha \nabla_{\Phi}{\hat{\mathcal{L}}_{\mathcal{T}_i}(f_\Phi)}})
\end{aligned}
\end{equation}

In the above equation $f_{\Phi}$ is the model, including the head and $\Phi$ are the entire model parameters. Note that since all prior meta learning approaches optimize on a set of tasks (subtasks as per our terminology), the loss function, unlike our notation, would be $\mathcal{L}_{\mathcal{T}_i}$. To avoid confusion, when we write (sub)tasks, it means others have considered them as tasks, but in our terminology, they are subtasks. Also, the actual notation of the model is $f_{\Phi}(\mathbf{I})$, where $\mathbf{I}$ is the input. As mentioned in section 2, \cite{RaghuRBV20} showed that Almost No Inner-Loop MAML (ANIL-MAML), a variant of MAML which only adapts the last layer, performs almost as well as MAML. Based on more elaborate experiments (section 5), we found that ANIL-MAML performs as well as MAML. Based on these advances, we use the ANIL-MAML training procedure:

\begin{equation}
\begin{aligned}
    \min_{\Phi, \theta} \sum_{\mathcal{T}_i \sim p(\mathcal{T})} \mathcal{L}_{\mathcal{T}_i}(g_{\theta^{'}}) = \sum_{\mathcal{T}_i \sim p(\mathcal{T})} \mathcal{L}_{\mathcal{T}_i}(g_{\theta - \alpha \nabla_{\theta}{\hat{\mathcal{L}}_{\mathcal{T}_i}(g_\theta)}})
\end{aligned}
\end{equation}

 In this case, $\Phi$ are the parameters of the model except the last layer or the head. In other words, if the model has $B$ conv-blocks, $\Phi$ represents the parameters of those conv-blocks and $\theta$ are the head parameters, which in most cases is a fully connected layer. Again, the actual notation for the head $g_{\theta}$ is $g_{\theta}(\mathbf{x})$, where $\mathbf{x} = f_{\Phi}(\mathbf{I})$, i.e. embedding or the output activations of $\mathbf{I}$ by the backbone network $f$. Note that, this notation of $g_{\theta - \alpha \nabla_{\theta}{\hat{\mathcal{L}}_{\mathcal{T}_i}(g_\theta)}}$ is valid for (sub)tasks which have task-shifts in them. We term this \textit{(sub)task adaptation}. For tasks whose (sub)tasks have only domain shift, we use \textit{domain adaptation} by pre-training (training all the data together without any task distinction; see Fig.~\ref{fig:compgraph} and refer to \cite{FinnAL17}) instead of meta-training. In which case $g_{\theta - \alpha \nabla_{\theta}{\hat{\mathcal{L}}_{\mathcal{T}_i}(g_\theta)}}$ is replaced by $g_{\theta}$.

To generalize the above training process to multiple high-level tasks of different dimension heads, instead of having a single head, we will now have a set of heads, each for solving a specific high level task. In other words, all task-specific heads, are meta-parameters and have an update rule mentioned in the previous equation. $\Phi$ would be then updated by summing over all the gradients obtained from the high-level tasks, as shown below.
 
\begin{equation}
\begin{aligned}
    \min_{\Phi} \    \sum_{\mathscr{T}_j} \     \sum_{\mathcal{T}_i^j} \mathcal{L}^j_i(g^j_{\theta^{'}}) = \sum_{\mathscr{T}_j} \sum_{\mathcal{T}^j_i} \mathcal{L}_i^j(g^j_{\theta - \alpha \nabla_{\theta}{\hat{\mathcal{L}}_i^j(g^j_\theta)}})
\end{aligned}
\end{equation}

\begin{figure*}
\begin{minipage}{.6\linewidth}
\includegraphics[width=\textwidth]{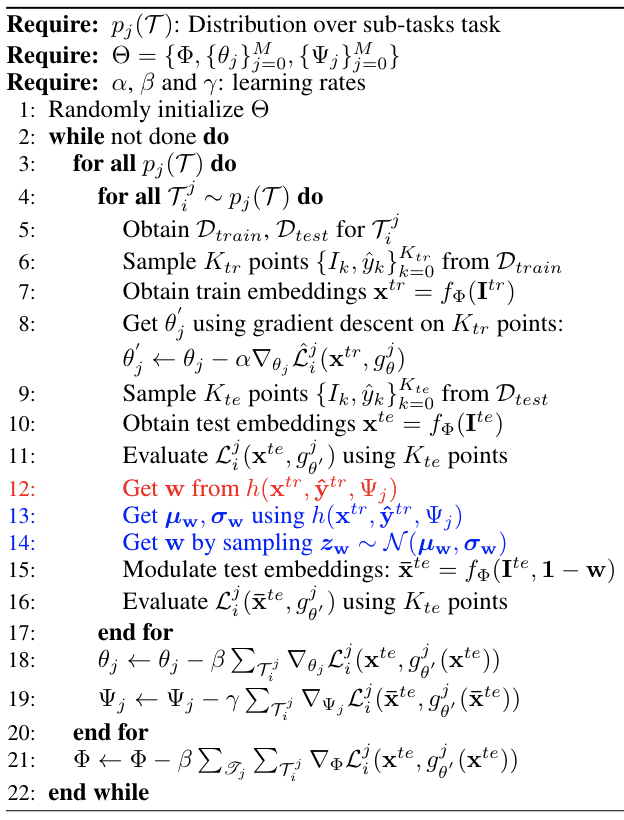}
\end{minipage}
\begin{minipage}{.4\linewidth}
\includegraphics[width=\textwidth, valign=t]{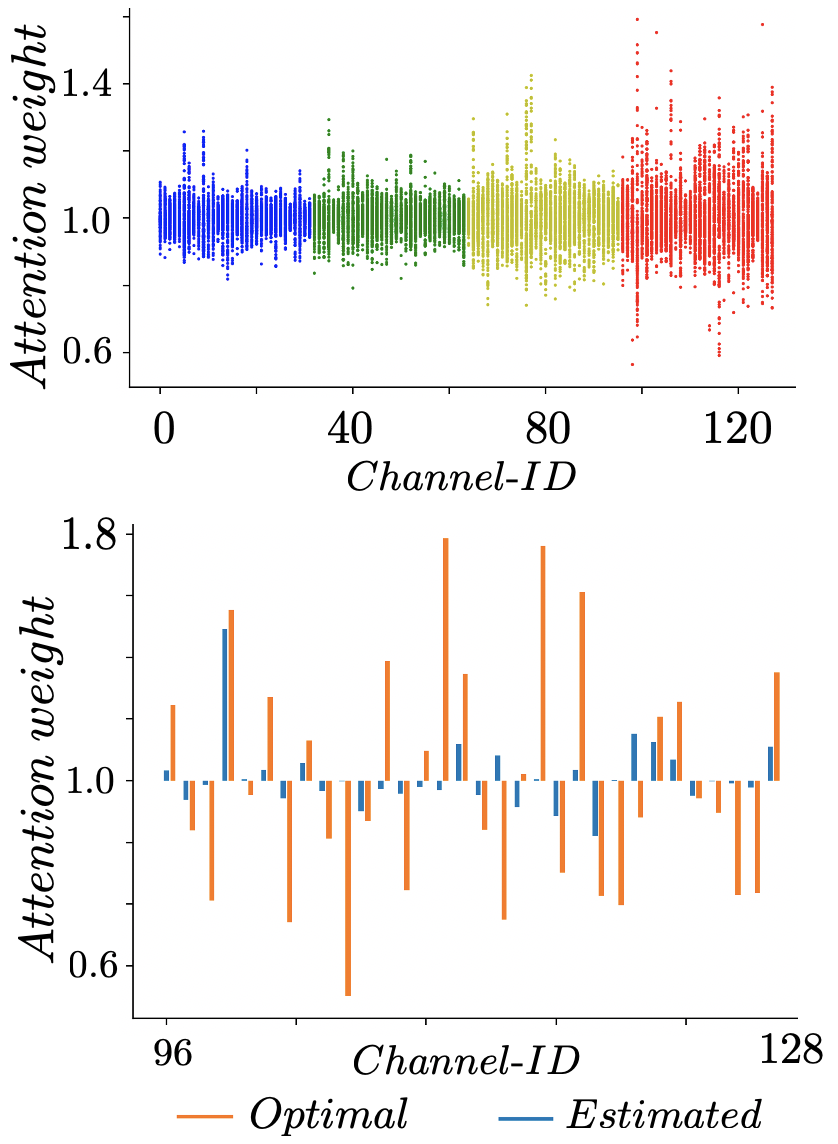}
\caption{Left: Algorithm. Top: Visualization of the attention weights obtained over 150 subtasks. 4 colors represent 4 conv-blocks. Bottom: Estimated and optimal attention weights for the last block. Most of the estimated weights point in the right direction of importance.}
\label{fig:weights}
\end{minipage}
\setlength{\belowcaptionskip}{-8pt}
\end{figure*}

\subsection{Attention Modulator}
To explain our attention approach, we further discuss the structure of our backbone network $f$. It is a Deep CNN with $B$ convolution blocks and $C$ channels per block. The total set of output activations obtained at end of each block for embedding input $\mathbf{x}$ are $\{o_i\}_{i=0}^{B*C}$, or $\mathbf{o}$, which are passed on to the next block except the last layer activations. As the importance of feature maps tends to change depending on task, we weight each by applying a channel-wise dot product, $\odot$, to each of the activations with attention weights $\mathbf{w}$ (initialized to ones). Mathematically: updated set of activations $\{\bar{o}_i\}_{i=0}^{B*C} = \mathbf{o} \odot (\boldsymbol{1} - \mathbf{w})$.

Our proposed general purpose task-based attention module $h_j$ parameterized by $\Psi_j$, learns to output these weights, by using the last layer activations $\mathbf{x}$ and labels $\mathbf{\hat{y}}$ of the train data of a specific task. Note that we also represent the flattened last layer activations $\{{o_i}\}_{i=(B-1)*C}^{B*C}$ as $\mathbf{x}$ and are passed to a task-specific head $g^j$. $h_j$ takes in the concatenation of pre-modulated input embeddings $\mathbf{x}$ of the input data $\mathbf{I}$ and their labels $\mathbf{\hat{y}}$ to output $\mathbf{w}$. This essentially forms a closed-loop modulation mechanism. The vector $\mathbf{w}$ is then used to weight the current meta feature-maps $\mathbf{o}$, as mentioned in the previous paragraph. The inputs $\mathbf{I}$ are again forwarded through the entire backbone network with the weighted feature-maps $\mathbf{\bar{o}}$ to obtain $\mathbf{\bar{x}}$ (Fig.~\ref{fig:arch}). The notation of this additional parameter is folded into $f$ as $f_{\Phi}(\mathbf{I}, \mathbf{1}-\mathbf{w})$. Initially, when the weights are $\mathbf{1}$, $f_{\Phi}(\mathbf{I})$ means the same as $f_{\Phi}(\mathbf{I}, \mathbf{1})$.

The attention network is modelled as a multi-\textit{attention}-layer network where each layer consists of an \textit{attention} block which transforms the input into Query, Key and Value pairs and computes dot-product attention as proposed in \cite{VaswaniSPUJGKP17}. Forward propagation through the attention network happens similar to a multi-layer perceptron, and the output of the first attention block is fed to the second and so on. Since each task inherently has different feature structures, the attention modulation for the low-level subtasks differs based on the high-level task. To overcome this, we use an attention module for each high-level task, while training multiple tasks together. See Appendix.~B for specific implementation details.

 \begin{figure}
  \includegraphics[width=\textwidth]{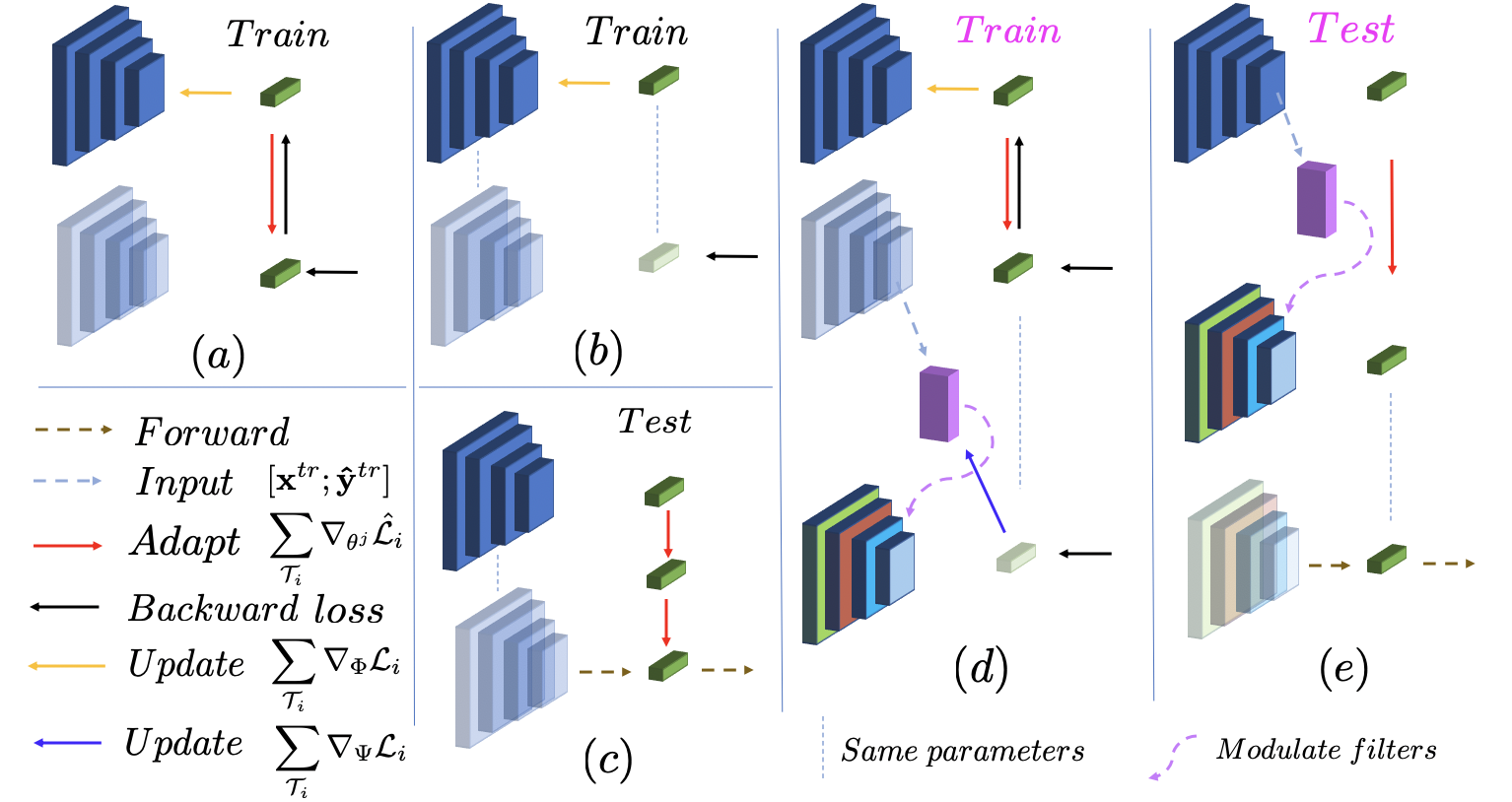}
  \caption{Computation graphs for the baseline model and our proposed method for each high-level task.  Dashed and bold arrows correspond to forward passing and gradient updates respectively which are higlighted for Training and Testing.  The shaded network indicate that the parameters are preserved during adaptation. Colored network represents the parameters after modulation. Green and violet modules are the task-specific head and the attention network respectively. (a) Training during \textit{Task adaptation}, (b) Training during \textit{Domain adaptation}, (c) Testing during \textit{Domain} and \textit{Task adaptation}, (d) Training and (e) Testing using attention module during \textit{Task adaptation}.}
  \label{fig:compgraph}
\end{figure}

We obtain the embeddings $\mathbf{x}$, by doing a forward pass of the backbone network. From the adapted task-specific heads, we then obtain gradients wrt the backbone parameters using the loss $\mathcal{L}^j_i$ on the test data. With the addition of the attention module $h_i$, the feature embeddings $\mathbf{x}$ are modified to $\mathbf{\bar{x}}$ by modulating the activations out of $f$. $\mathbf{\bar{x}}$ is now, passed to the task-specific head $g^j$, to obtain the output, which are then used to obtain gradients wrt parameters $\Psi$ using $\mathcal{L}^j_i$. $f$ and $h_i$ are optimized this way. The update rules for $\Phi$ and $\Psi_j$ are given below. Computation graphs of \textit{task adaptation}, \textit{domain adaptation} along with the attention-module are shown in Fig.~\ref{fig:compgraph}.

\begin{equation}
    \Phi \leftarrow \Phi - \beta \sum_{\mathscr{T}_j} \sum_{\mathcal{T}_i^j} \nabla_{\Phi} \mathcal{L}^{j}_{i}(f_{\Phi}(\mathbf{I}), g^j_{\theta^{'}})
\end{equation}
\begin{equation}
    \Psi_j \leftarrow \Psi_j - \gamma \sum_{\mathcal{T}_i^j} \nabla_{\Psi_j} \mathcal{L}^{j}_{i}(f_{\Phi}(\mathbf{I}, \boldsymbol{1}-\mathbf{w}), g^j_{\theta^{'}})
\end{equation}

\paragraph*{Probabilistic Attention.} Since there is inherent uncertainty in the problem of few-shot adaptation, we could also model the attention module to output a parametric distribution of attention weights rather than a point estimate $\mathbf{x}$. We model the distribution as a Gaussian, whose parameters are obtained as an output from the attention module. The loss function, to optimize $\Psi_j$ in this case, is the Evidence Lower bound (ELBO):

\begin{align*}
    \log(\mathbf{y}|\mathbf{I}^{te}, \mathbf{I}^{tr},\mathbf{\hat{y}}^{tr}) \geq \mathbb{E}_{q(\boldsymbol{z_\mathbf{w}}|\mathbf{x}^{te}, \mathbf{\hat{y}}^{te})} \left[\mathcal{L}(f_{\Phi}(\mathbf{I}^{te}, \mathbf{1}-\boldsymbol{z_\mathbf{w}}), g_{\theta^{'}}) + KL[p \;\boldsymbol{||}\;  q] \right]
\end{align*}

Note that the indices for high-level and low-level tasks have been omitted for simplicity. Following a similar approach to \cite{abs-1807-01622}, we model prior $p(\boldsymbol{z_\mathbf{w}}|\mathbf{x}^{tr}, \mathbf{\hat{y}}^{tr})$ and posterior $q(\boldsymbol{z_\mathbf{w}}|\mathbf{x}^{te}, \mathbf{\hat{y}}^{te})$ distributions as $\mathcal{N}(\boldsymbol{\mu_\mathbf{w}}, \boldsymbol{\sigma_\mathbf{w}})$ respectively. We approximate $q$ in place of $p$, during train-time and minimize the KL divergence of both the distributions along with minimizing the test loss. Since the posterior is not available during testing, as we do not have access to the labels of the test data, we use the prior $p$. Our final method is summarized in Fig.~\ref{fig:weights}, with both variants of the attention module highlighted in different colors.

 \begin{figure*}
 \includegraphics[width=\textwidth]{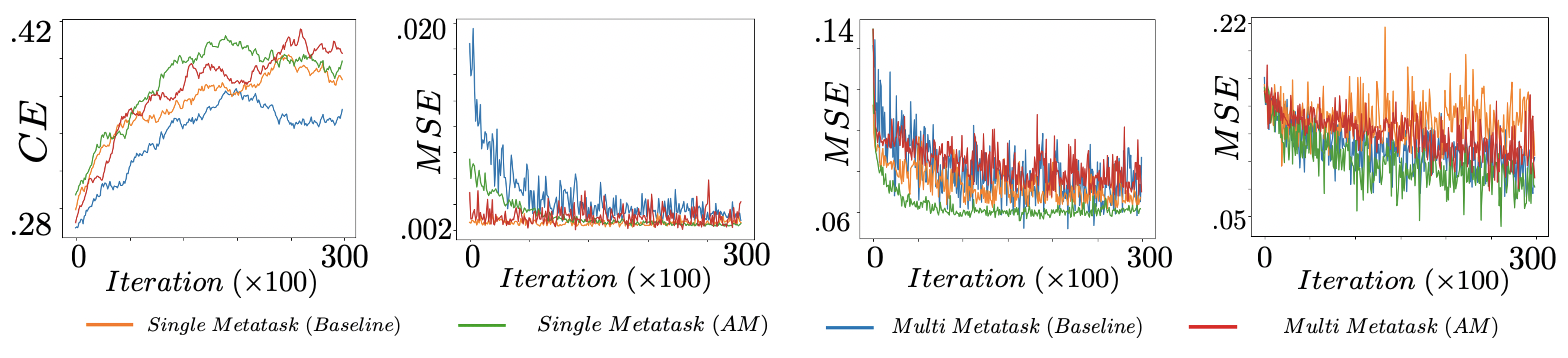}

  \caption{Validation curves for each high-level task (order as mentioned in Sec.5) in the MMT dataset using the baselines presented in Table~\ref{tab:mmt}. Our method achieves better accuracy or lower loss compared to other baselines. MSE and CE stands for mean square error and cross-entropy respectively.}
  \label{fig:curves}
\end{figure*}

\section{Experiments}
\textbf{Datasets and Task details.} To evaluate our model, we propose the Multi Meta Tasks dataset (MMT), a dataset of datasets \cite{ZhouLKO018, VinyalsBLKW16, DaiCSHFN17, abs-1908-00463, SilbermanHKF12} with 4 high-level tasks: Scene classification, depth-estimation, surface-normal estimation and vanishing point estimation. Scene classification is a meta-supervised problem solved using \textit{task-adaptation} (different output labels for each subtask), whereas the other tasks are \textit{domain adaptation} problems. Unlike many other multi-task evaluation benchmarks, for each high-level task, train and test datasets are different (Appendix A).

\textbf{Task Architectures.} The backbone network is a CNN with 4 conv-blocks, each of 32 filters of 3x3 convolutions with batch-norm and max-pooling. We used fully connected layers as heads for Scene Classification and Vanishing point estimation. As for Depth and Surface Normal estimation, the heads are 4 block conv-transpose blocks, with 4 and 8 filters in each block, respectively. A conv-transpose block consists of a conv-transpose layer followed by a convolution layer. For tasks having labels in the pixel domain, like that of depth and surface-normal estimation, we additionally use a 3-block CNN with 4 filters in each block, to compress the label to vector space and concatenate with feature embedding $\mathbf{x}$. Because of lack of space, we mention the additional architecture and training details and hyperparameters in Appendix B.

\subsection{Evaluation on MMT Dataset}
Specifically, we compare our method with multiple baselines using the MMT dataset. \textit{Single MetaTask learning}, where meta learning is performed only one a specific high-level task, \textit{Multi MetaTask learning}, where multiple high-level tasks are meta-learnt using a common representation network. For each of these baselines, we use our attention module to show the differences. We trained our model on all the tasks in every iteration. Each high-level task has a standalone attention module. As loss functions, mean square error is used for all tasks except scene classification, which uses cross-entropy. Results are in Table \ref{tab:mmt} and Fig.~\ref{fig:curves}. Similar to prior works on multi-task learning, we also faced challenges with balancing the loss functions of individual tasks, as some tasks overfit. In the current work, we manually hard-coded fixed weights for the loss functions of each high-level task. However, we believe that dynamically estimating weights of each loss function may lead to better performance in future work.

\subsection{Meta Learning for Image Classification}
To compare our method with other meta learning methods, we assess our model on mini-imagenet (Table~\ref{tab:ic}). Although recent works use better feature extraction networks \cite{MishraR0A18}, we stick with the standard 4-Conv network as used in the original work \cite{FinnAL17}. We also use the NIL metric as proposed in \cite{RaghuRBV20} to assess the quality of embeddings learnt by our method. NIL involves having no head, and class labels of the test-set are determined by cosine distance from the samples in the train-set. We compare our methods with baselines MAML, ANIL-MAML on mini-places, a mini-imagenet variant of the Places-365 dataset \cite{ZhouLKO018} (Appendix A), after training on mini-imagenet. Our method beats all but one method using a similar architecture, with a training and inference speed-up of 1.7x and 2.3x relative to MAML and using just 1.1x more parameters than MAML. Note that none of the compared methods (including the one that performs better than ours) can handle multiple high-level tasks, which is the main strength of our new approach.

\subsection{Visualizing Attention weights}
We conducted experiments to understand how optimal our attention weights are. We took a network meta-trained on mini-imagenet and initialized all the channel weights to $\boldsymbol{1}$. We then select a test task and train these weights, alone, with all the other parameters fixed (including the adapted head), using all the data available for that specific task. These final weights are compared with the predicted attention weights (Fig.~\ref{fig:weights}). Overall, we find that, although the magnitude of these weights are not equal to the optimal ones, they point in the right direction of importance.


\begin{table*}[ht]
\small
\centering
\caption{Comparison of the proposed method with the baselines evaluated on the MMT dataset. Accuracy is given for classification along with 95\% confidence intervals over all the subtasks. Accuracy mentioned for depth, surface normal and vanishing point is the mean percentage of the output within the threshold of $10^{-3}$ of the label. Error mentioned in this table denotes mean squared error. Error for Depth is of the order $\times 10^{-3}$.}
\begin{center}
\begin{tabular}{lllllllll}
\Xhline{2\arrayrulewidth}
 Tasks & Filters & \multicolumn{1}{c}{\small{Classification}} &  \multicolumn{2}{c}{Depth} & \multicolumn{2}{c}{VP} & \multicolumn{2}{c}{Normal} \\
 \Xhline{2\arrayrulewidth}
  &  &  Accuracy   & Acc & Error   & Acc &  Error   & Acc & Error\\
 \hline
 \multirow{2}{*}{\makecell{Single meta task \\ (Baseline)}} & 32 & 37.7 ${\pm}$ 0.35 & 86.5 & 0.543 & 10.5 & 0.1058 & 8.9 & 0.0936\\
 & 64 & 38.42 ${\pm}$ 0.35 & 81.4 & 0.384 & 11.1 & 0.103 & 12.03 & 0.095\\ \hline

  \multirow{2}{*}{\makecell{Single meta task \\ (AM)}} & 32 & 39.6 ${\pm}$ 0.41 & \textbf{87.8} & 0.495 & \textbf{17.6} & 0.0823 & 11.4 & 0.0841\\
& 64 & 39.45 ${\pm}$ 0.45 & 88.6 & \textbf{0.361} & 16.8 & \textbf{0.071} & 14.5 & 0.082\\ \Xhline{2\arrayrulewidth}

 %
  \multirow{2}{*}{\makecell{Multi meta task \\ (Baseline)}} & 32 & 37.88 ${\pm}$ 0.43 & 80.71 & 0.698 & 11.3 & 0.1183 & 9.4 & 0.0842\\
& 64 & 39.94 ${\pm}$ 0.39 & 76.1 & 0.867 & 13.8 & 0.103 & 12.5 & 0.066\\ \hline
 
 %
  \multirow{2}{*}{\makecell{Multi meta task \\ (AM)}} & 32 & 39.07 ${\pm}$ 0.42 & 78.06 & 0.852 & 17.4 & 0.081 & \textbf{13.6} & 0.0621\\
& 64 & \textbf{39.99 ${\pm}$ 0.41} & 85.3 & 0.527 & 16.2 & 0.088 & 12.5 & \textbf{0.0613}\\ \hline
 
 %

\hline
\end{tabular}
\end{center}
\label{tab:mmt}
\end{table*}%

\begin{table*}[ht]

\small
\caption{Left: Quantitative comparison of our method with other state of the art meta learning methods on 5-way, 1-shot and 5-shot classification tasks from mini-imagenet. $\pm$ shows 95\% confidence intervals. Right: To ascertain the quality of the embeddings, we also use NIL metric apart from the standard cross-entropy (CE) loss. Evaluation is performed on mini-places (mP) and mini-imagenet (mI), after the model is trained only on mini-imagenet. Results given are for 5-way 1-shot (sub)tasks.}
\begin{minipage}{.5\linewidth}
\begin{tabular}{llll}
\Xhline{2\arrayrulewidth}
Method  & Backbone  & 1 shot   &   5 shot \\
\Xhline{2\arrayrulewidth}
 MetaLearner \cite{RaviL17}  & Conv-4   & 43.44 ${\pm}$ 0.77    & 60.60 ${\pm}$ 0.71\\
 MatchingNet \cite{VinyalsBLKW16}  & Conv-4   & 43.56 ${\pm}$ 0.84    & 55.31 ${\pm}$ 0.73\\
 ANIL \cite{RaghuRBV20}  & Conv-4   & 48.1 ${\pm}$ 1.51     & 61.0 ${\pm}$ 0.6\\
 MAML \cite{FinnAL17}  & Conv-4    & 48.7 ${\pm}$ 1.84    & 63.1 ${\pm}$ 0.4\\
 BMAML \cite{YoonKDKBA18} & Conv-4    & \textbf{53.8 ${\pm}$ 1.46}         & -\\
 EMAML \cite{YoonKDKBA18} & Conv-4  & 51.04 ${\pm}$ 1.46         & -\\
 AM (Ours) & Conv-4 & 51.1 ${\pm}$ 0.23          & \textbf{64.6 ${\pm}$ 0.52} \\
 P-AM (Ours) & Conv-4 & 48.9 ${\pm}$ 0.94          & 63.17 ${\pm}$ 0.11\\
 
 \hline
 
\hline
\end{tabular}
\end{minipage}
\begin{minipage}{.6\linewidth}
\centering
\small
\begin{tabular}{llll}
\Xhline{2\arrayrulewidth}
Method  & CE & NIL \\
\Xhline{2\arrayrulewidth}
 MAML (mI)    & 48.7        & 49.1 \\
 MAML (mP)    & 31.2        & 33.8 \\
 \hline
 ANIL (mI)   & 48.1       & 49.5 \\
 ANIL (mP)  & 32.7        & 33.9 \\
 \hline
 AM (mI)   & \textbf{51.1}        & \textbf{52.3} \\
 AM (mP)   & \textbf{33.5}         & \textbf{34.3} \\
 \hline
 P-AM (mI)   & 48.9        & 50.2 \\
 P-AM (mP)   & 32.6         & 34.1 \\
\Xhline{2\arrayrulewidth}

\end{tabular}
\end{minipage}
\label{tab:ic}
\end{table*}%

\section{Conclusion}
We formulated a multi-task meta learning problem where a single model needs to execute multiple heterogeneous tasks. The core of this problem was to learn task-invariant representations, apart from learning meta-parameters of the head for each task. These heads are used to adapt to unseen subtask belonging within each high-level task. As a baseline, we modified the MAML framework by including multiple heads and adopting ANIL training. We then presented a flexible attention mechanism, which could be applied in a wide variety of \textit{task} or \textit{domain adaptation} scenarios. This makes the adaptation better by providing inductive bias on what features to focus onto.

We also contributed a multi-metatask dataset, a dataset of datasets of different high-level tasks, for evaluation. Results using this dataset highlight the improvements of our proposed method in the presented scenario. Lastly, we also showed performance gains compared to existing meta learning algorithms, on mini-imagenet.

During experimentation, we faced many challenges in regards to loss balancing, as some high-level tasks either had more data or faster gradient updates, which suffocated other tasks. Future work could involve loss balancing strategies for seamless meta learning of multiple modalities together.

\section{Broader Impact}

Deep learning has recently surpassed traditional approaches in many real world problems. By merely having sufficient data, we can develop a system which can perform a task, sometimes nearly flawlessly. Some of the most popular applications, related to computer vision are Facial Recognition, Autonomous driving, Object detection, etc.

Our work, when packaged into an adaptive system, could be deployed by using data on the fly. We augment the abilities of deep-learning by enabling systems to adapt and infer, rather than only inferring at test-time. This allows the system to be more robust, as the system will gain flexibility by learning only task-invariant priors as the actual performance data would be provided during test-time. Also, instead of training a model from scratch for every new application, we could use the current meta-trained system as a warm start. In robotics, these systems could also be used for scene understanding, which helps the robotic agents determine their state.



\section{Acknowledgements}
This work was supported by the National Science Foundation (grants CCF-1317433 and CNS-1545089) and Intel Corporation. The authors affirm that the views expressed herein are solely their own, and do not represent the views of the United States government or any agency thereof.

\bibliographystyle{plainnat}
\bibliography{main}

\end{document}